\documentclass[letterpaper, 10 pt, conference]{ieeeconf}

\usepackage{graphicx}
\usepackage{subcaption}
\usepackage{caption}
\usepackage{nicefrac}       
\usepackage{array}
\usepackage{float}
\usepackage{microtype}
\usepackage{graphicx}
\usepackage{amsmath}
\usepackage{amssymb}
\usepackage{booktabs} 
\usepackage{bbm}
\usepackage{multirow}
\usepackage{multicol}
\usepackage{hhline}
\usepackage{algorithm}
\usepackage{algorithmic}
\usepackage{hyperref}
\usepackage{cleveref}

\makeatletter
\newcommand{\thickhline}{%
    \noalign {\ifnum 0=`}\fi \hrule height 1pt
    \futurelet \reserved@a \@xhline
}
\newcolumntype{"}{@{\hskip\tabcolsep\vrule width 1pt\hskip\tabcolsep}}
\makeatother

\makeatletter
\newcommand\fs@spaceruled{\def\@fs@cfont{\bfseries}\let\@fs@capt\floatc@ruled
  \def\@fs@pre{\vspace{1\baselineskip}\hrule height.8pt depth0pt \kern2pt}%
  \def\@fs@post{\kern2pt\hrule\relax}%
  \def\@fs@mid{\kern2pt\hrule\kern2pt}%
  \let\@fs@iftopcapt\iftrue}
\makeatother

\IEEEoverridecommandlockouts                              

\DeclareMathOperator*{\argmax}{arg\,max}

\title{\LARGE \bf
Bayesian Curiosity for Efficient Exploration in Reinforcement Learning
}

\author{Tom Blau$^{1}$, Lionel Ott$^{1}$, Fabio Ramos$^{1,2}$
\thanks{* Correspondence to: Tom Blau, {\tt\small tom.blau@sydney.edu.au}}%
\thanks{$^{1}$ School of Computer Science, The University of Sydney, Australia}%
\thanks{$^{2}$ NVIDIA, USA}%
}

\begin{document}
\maketitle
\global\csname @topnum\endcsname 0
\global\csname @botnum\endcsname 0

\begin{abstract}
Balancing exploration and exploitation is a fundamental part of reinforcement learning, yet most state-of-the-art algorithms use a naive exploration protocol like $\epsilon$-greedy. This contributes to the problem of high sample complexity, as the algorithm wastes effort by repeatedly visiting parts of the state space that have already been explored. We introduce a novel method based on Bayesian linear regression and latent space embedding to generate an intrinsic reward signal that encourages the learning agent to seek out unexplored parts of the state space. This method is computationally efficient, simple to implement, and can extend any state-of-the-art reinforcement learning algorithm. We evaluate the method on a range of algorithms and challenging control tasks, on  both simulated and physical robots, demonstrating how the proposed method can significantly improve sample complexity.
\end{abstract}

\section{Introduction}\label{intro}
\begin{figure}[t]
\begin{center}
\centerline{\includegraphics[width=\columnwidth]{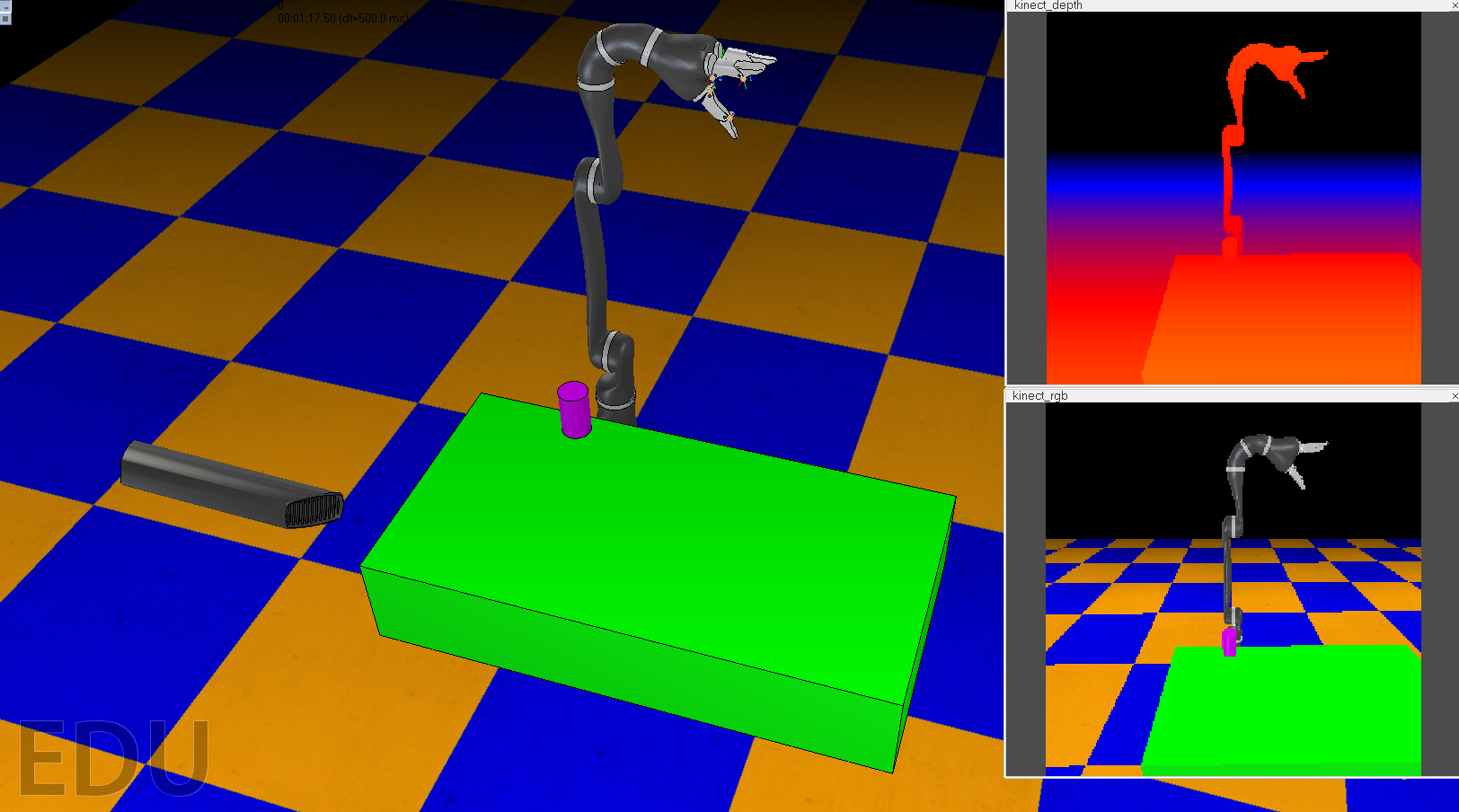}}
\caption{Robotic reaching/grasping environment. On the left is a plain camera view of the environment, including the visual sensor. On the top right, the depth channel of the sensor displayed as a heatmap. On the bottom right, the RGB channels of the sensor.}
\label{fig:grasp-sim}
\end{center}
\vskip -0.2in
\end{figure}
Reinforcement Learning is a powerful approach for learning control policies for tasks that require a sequence of decisions, such as walking, playing a game, or navigating a maze. This learning is accomplished by repeatedly taking actions and observing both their effect on the environment state and some scalar reward signal that they elicit. The parameters of the policy are iteratively updated to increase the total expected reward over an entire sequence of observations and actions, called a trajectory. The weakness of RL is that it suffers from high sample complexity. Often millions of actions must be taken before an RL algorithm learns an efficient control policy, meaning a single execution can take days to complete. This problem is particularly pronounced in tasks where the space of possible actions and states is continuous and high-dimensional, a category that includes most real-world tasks, such as robotic object manipulation or autonomous driving.

One contributor to this high sample complexity is the exploration-exploitation trade-off~\cite{sutton1998reinforcement}. This is a fundamental trade-off in any sequential decision making problem between \textit{exploitation} - using available knowledge to follow a trajectory that maximizes expected reward - and \textit{exploration} - searching previously unseen parts of the state space to find better trajectories. Excessive exploitation slows learning as effort is spent achieving high rewards without gaining new information that can be used to improve the policy. On the other hand, excessive exploration slows learning as effort is spent searching through regions of the state space where improved trajectories are unlikely to be found. Achieving good exploration is a challenging problem, as all but the simplest RL tasks have state spaces that are intractable to explore fully. The state space grows exponentially in the number of dimensions, and the relationship between actions and state changes is often complex, so that even reaching a specific state is not trivial.

In spite of this, it is common for state-of-the-art RL algorithms to handle exploration in a naive manner. A standard approach is $\epsilon$-greedy, which consists of taking a random action with probability $\epsilon$ and following the learned policy with probability $1-\epsilon$. The equivalent for continuous action spaces is adding Gaussian noise to policy actions. Consequently, exploration is often inefficient, spending much time searching through well-known regions of the state space that will provide little new information, or through regions that good trajectories are unlikely to intersect.

More recently, a class of algorithms has emerged that guides learning by use of intrinsic rewards~\cite{oudeyer2008can}, meaning a reward signal that comes from the model itself (intrinsic) rather than the environment (extrinsic). This is analogous to curiosity in humans, which are internally motivated to seek out novel experiences, even when there is no external reward for doing so. However, such algorithms tend to require cumbersome additions to the policy model.

In this paper we propose a curiosity based method that can be applied to an arbitrary RL algorithm in order to direct exploration towards parts of the state space that have not yet been visited. Using a Bayesian linear regression model with a learned latent space embedding, we compute the uncertainty of the model for arbitrary states. This uncertainty grows with the dissimilarity from previously seen points in the state space, and is therefore a good measure of the novelty of a given state observation. We use the model uncertainty to generate an intrinsic reward signal that encourages visiting new states far from previously explored regions. Applying this approach to state-of-the-art RL algorithms, we verify experimentally that it accelerates learning in classic control tasks as well as in challenging robotics tasks with high dimensional state spaces, continuous action spaces, and even sparse rewards.


\section{Related Work}\label{related}
Trading off exploration and exploitation is a fundamental problem in reinforcement learning, dating back to the multi-armed bandit problem~\cite{robbins1985some} studied in statistics. The canonical algorithm for handling this trade-off was $\epsilon$-greedy~\cite{sutton1998reinforcement}, which consists of taking a random action with probability $\epsilon$ and otherwise taking a greedy action that maximizes the expected reward. Although guaranteed to converge, $\epsilon$-greedy is unable to incorporate information about the model's uncertainty. Many algorithms have been proposed to leverage such uncertainty information by estimating a posterior over the expected value, for example using Bayesian linear regression~\cite{rlsvi,2018bdqn} or Bayesian neural networks~\cite{lipton2018bbqn}. While provably efficient, these algorithms involve the use of an argmax operation, limiting them to problems with discrete actions. Our method seeks to address exploration in more realistic problems with continuous actions.

One powerful approach that can use uncertainty information to deal with the exploration-exploitation trade-off is the Upper Confidence Bound (UCB) algorithm~\cite{ucb}. Instead of greedily taking the action that maximizes the expected reward, the UCB algorithm takes the action that maximizes the sum of the reward's expectation and variance. A significant weakness, however, is that UCB does not extend naturally beyond the bandit setting. The EMU-Q algorithm~\cite{morere18a} applies UCB principles to more general reinforcement learning problems. Bayesian linear regression with random Fourier features is used to estimate the Q-function as well as the uncertainty of the model and actions are taken that greedily maximize the sum of both. An alternative approach is SARSA with an uncertainty Bellman equation (UBE)~\cite{ube}. The authors derive a different Bellman backup for the uncertainty of the Q-function, and approximate a Bayesian linear regression with a DQN~\cite{mnih2015human}. Actions are taken by Thompson sampling~\cite{thompson1933}. While the EMU-Q algorithm works well for continuous action spaces with low-dimensional actions and observations, UBE works well in problems with high-dimensional observations but is restricted to discrete action spaces. In contrast, the method we propose can solve problems that have both a continuous action space and high-dimensional observations such as images.

In recent years, intrinsic motivation algorithms have emerged as a popular solution to the exploration-exploitation trade-off. This class of algorithms is based on the idea that exploration can be driven by a separate reward signal that encourages visiting under-explored states. Such a reward signal can be derived from visitation counts~\cite{bellemare2016,machado2018count,tang2017exploration}, model prediction error~\cite{burda2019exploration, pathak2017curiosity,achiam2017surprise}, variational information gain~\cite{houthooft2016vime}, or entropy maximization~\cite{eysenbach2018diversit,hong2018diversity}. In contrast, we propose to use the uncertainty of a Bayesian linear regression, which is a well understood mathematical mechanism that can explicitly separate uncertainty in the model from uncertainty in the data. In~\cite{bechtle2019curious}, the uncertainty of a Gaussian process was used as a reward term in a linear-quadratic regularizer in order to encourage exploration. While Gaussian processes provide good uncertainty estimates, they are problematic when it comes to dealing with large data sets and high-dimensional data such as images, both of which our method can handle.

\section{Background}\label{background}
\subsection{The Markov Decision Process}
A Markov Decision Process (MDP) is described by a tuple $(\mathcal{S}, \mathcal{A}, \pi, \mathcal{P}, r, \gamma, H)$ where:
$\mathcal{S}$ is the set of all possible states. $\mathcal{A}$ is the set of all possible actions. $\pi : \mathcal{S} \times \mathcal{A} \rightarrow [0, 1]$ is a stochastic policy mapping state-action pairs $(s, a)$ to the probability of choosing action $a \in \mathcal{A}$ in state $s \in \mathcal{S}$. $\mathcal{P} : \mathcal{S} \times \mathcal{A} \times \mathcal{S} \rightarrow [0, 1]$ is a state transition function mapping tuples $(s_{t}, a_{t}, s_{t+1})$ to the probability of arriving at state $s_{t+1}$ after taking action $a_{t}$ at state $s_{t}$. $r : \mathcal{S} \times \mathcal{A} \rightarrow \mathbb{R}$ is a reward function assigning a scalar reward value to each state-action pair. $\gamma \in (0, 1)$ is a discount factor while $H \in \mathbb{N}$ is a time horizon.

An MDP can be used to generate a sequences of states and actions as follows: given an initial state $s_0$, iteratively select the next action $a_t \sim \pi(s_t)$ and evolve the state by sampling $s_{t+1} \sim \mathcal{P}(s_t, a_t)$. The sequence $[s_0, a_0, \dots, a_{H-2}, s_{H-1}]$ is called a \textit{trajectory}. The expected total discounted reward over an entire trajectory is:
\begin{equation}
    J_\pi = \mathbf{E}_\pi\left[\sum_{t=0}^{H-1} \gamma^tr(s_t, a_t)  \right],
\end{equation}
where the expectation is subscripted by $\pi$ to denote the distribution over trajectories that $\pi$ induces, and $\gamma$ discounts rewards later in the trajectory. Let $\pi_{\theta}$ denote a parameterized policy whose parameters are $\theta$. A reinforcement learning algorithm seeks to maximize the expected total discounted reward by optimizing over the policy parameters. This can be expressed succinctly as the following optimization problem:
\begin{equation}
    \theta^* = \argmax_\theta J_{\pi_\theta}.
\end{equation}

\subsection{Bayesian Linear Regression}\label{blr}
Linear regression is a form of regression analysis where the data is explained using a linear model~\cite{yan2009linear}. That is, a model of the form $t = \mathbf{w} \cdot \mathbf{x} + b$, where $\mathbf{x}$ is the input, $\mathbf{w}$ is a vector of weights and $b$ is the intercept. Bayesian linear regression (BLR)~\cite{Bishop2006} places a prior on the model weights $\mathbf{w}$. We choose a Gaussian prior:
\begin{equation}
\begin{aligned}
	&p(\mathbf{w}) = \mathcal{N}(\mathbf{w}|\mathbf{m}_{0}, \mathbf{S}_{0}), \\
	&\mathbf{m}_0 = \mathbf{0} \quad\text{and}\quad
	\mathbf{S}_0 = \alpha^{-1}\mathbf{I},
\label{eq:prior}
\end{aligned}
\end{equation}
where $\alpha$ is a hyperparameter representing the precision of the model, and $\mathbf{m_0}, \mathbf{S_0}$ are a vector and matrix describing the prior mean and covariance over the model weights. Note that in order to simplify the notation we remove the intercept $b$ and instead add a dummy dimension to the input $\mathbf{x}$ that will always have a constant value of 1. Let there be a set of training data ${(\mathbf{X}, \mathbf{t})}$  where $\mathbf{X}$ is an $N\times{}D$ matrix whose rows are $D$-dimensional input vectors and $\mathbf{t}$ is an $N$-dimensional vector of targets. Although we restrict the targets to being scalar in order to simplify the notation, the following treatment can be easily extended to vector targets. In order to capture nonlinearity in the data, we transform the inputs of the BLR using a nonlinear function $\phi(x): \mathbb{R}^D\mapsto\mathbb{R}^M$. Note that it is possible that $M \neq D$, meaning $\phi$ can change the dimensionality of the data. The posterior distribution over $\mathbf{w}$ given the training data is:
\begin{equation}
\begin{aligned}
	&p(\mathbf{w}) = \mathcal{N}(\mathbf{w}|\mathbf{m}_{N}, \mathbf{S}_{N}),\\
	&\mathbf{m}_N = \mathbf{S}_N(\mathbf{S}_0^{-1}\mathbf{m}_0 + \beta\mathbf{\Phi}^T\mathbf{t}),\\
	&\mathbf{S}_N = \mathbf{S}_0^{-1} + \beta\mathbf{\Phi}^T\mathbf{\Phi},
\end{aligned}
\label{eq:posterior}
\end{equation}
where $\beta$ is a hyperparameter representing noise precision in the data, $\mathbf{t}$ is a vector of targets and $\mathbf{\Phi}$ is an $N\times{}M$ matrix whose $i$-th row is $\phi(x_i)$. Note that after we perform an update with one set of data, we can perform another update with a second set by assigning $\mathbf{m}_0 = \mathbf{m}_N; \mathbf{S}_0 = \mathbf{S}_N$.

Linear regression gives a prediction of $t$ for an arbitrary input $\mathbf{x}$. BLR with the above formulation yields a posterior predictive distribution over $t$:
\begin{equation}
\begin{aligned}
	&p(t|\mathbf{x}) = \mathcal{N}(t|\mathbf{m}_N^T\mathbf{\phi}(\mathbf{x}), \sigma_N^2(\mathbf{x})),\\
	&\sigma_N^2(\mathbf{x}) = \beta^{-1} + \mathbf{\phi}(\mathbf{x})^T \mathbf{S}_N\mathbf{\phi}(\mathbf{x}),
\end{aligned}
\label{eq:predictive}
\end{equation}
where $\sigma_N^2(\mathbf{x})$ is the variance of the predictive posterior.

\section{Bayesian Curiosity}\label{method}
\begin{figure}[b]
    \centering
    \includegraphics[width=\linewidth]{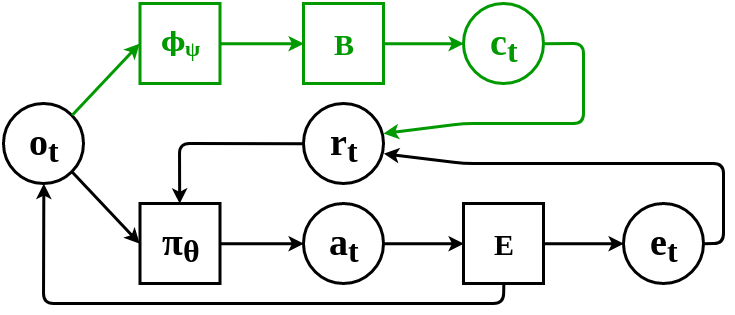}
    \caption{Pipeline of RL with Bayesian Curiosity. Squares are functions and circles are variables.}
    \label{fig:pipeline}
\end{figure}
We propose \textit{Bayesian Curiosity}, a method that can extend any RL algorithm by adding a secondary model that produces an intrinsic reward signal. The pipeline of the method can be seen in~\cref{fig:pipeline}. In black is the standard RL machinery: a state observation $o_t$ is passed to the policy network $\pi_\theta$, which produces action $a_t$. This action is then passed to the environment $E$, which evolves its state and emits both an environment reward $e_t$ and the next state observation. In green is the proposed Bayesian Curiosity mechanism: $o_t$ is given as input to a neural network $\phi_\psi$ with parameters $\psi$, which transforms it into a latent space. This latent space representation is passed to a BLR $B$, which computes the uncertainty for the given input. The uncertainty is then used to construct a curiosity reward signal $c_t$, which together with $e_t$ forms the reinforcement reward $r_t$.

\subsection{Latent Space Embedding}

\begin{algorithm}[b]
   \caption{Learning Latent Embedding}
   \label{alg:latent-embedding}
\begin{algorithmic}
   \STATE {\bfseries Input:} Training data $D$
   \STATE Randomly initialize $\phi_\psi$
   \STATE Initialize $\mathbf{m}_0, \mathbf{S}_0$ according to~\eqref{eq:prior}
   \REPEAT
   \STATE Randomly select subset $E \subseteq D$
   \STATE Randomly partition $D$ into minibatches $MB$
   \FOR{$mb \in MB$}
   \STATE Compute NLL for $mb$ and $E$ according to~\eqref{eq:loss}
   \STATE Take SGD step w.r.t. $\psi$ to minimize NLL
   \ENDFOR
   \UNTIL{improvement on the NLL loss has converged}
\end{algorithmic}
\end{algorithm}

\begin{algorithm}[b]
   \caption{Bayesian Curiosity}
   \label{alg:bayesian-curiosity}
\begin{algorithmic}
   \STATE {\bfseries Input:} Environment $\mathcal{E}$, time horizon $H$, RL algorithm $A$, learned transformation $\phi_\psi$, untrained BLR $B$
   \STATE Randomly initialize $\pi_\theta$
   \STATE Initialize $\mathbf{S}_0$ according to~\eqref{eq:prior}
   \REPEAT
   \STATE Reset $\mathcal{E}$ and observe $o_0$
   \STATE Create empty buffer $Z$
   \FOR{$i=1$ {\bfseries to} $H$}
   \STATE $a_{i-1}$ = $\pi_\theta(o_{i-1})$
   \STATE Execute $a_{i-1}$ and observe $o_i,e_i$ 
   \STATE Compute $r_i$ according to~\eqref{eq:curiosity} and~\eqref{eq:total}
   \STATE Add $(a_{i-1}, o_{i-1}, r_i)$ to $Z$
   \ENDFOR
   \STATE Update $\mathbf{S}_0$ according to \eqref{eq:posterior} using $\phi_\psi(o_{0:H-1})$
   \STATE Update $\pi_\theta$ using buffer $Z$ and algorithm $A$
   \UNTIL{policy improvement has converged}
\end{algorithmic}
\end{algorithm}

BLR is a linear model, and requires some transformation $\phi$ on the inputs in order to capture nonlinearity in the data. Further, it scales poorly with the dimensionality of the data. We will therefore learn a transformation using a neural network $\phi_\psi$ with weights $\psi$ to capture nonlinearity and reduce dimensionality.~\Cref{alg:latent-embedding} introduces a procedure for jointly optimizing the parameters of the network $\phi_\psi$ and BLR $B$, given a set of training data $D$. We acquire the training data by generating expert demonstrations with small Gaussian noise, meaning that our method is a form of learning from demonstration. In terms of the BLR, this means that $\mathbf{x}$ will be observations and $\mathbf{t}$ will be actions.

To train $\phi_\psi$ we follow a stochastic gradient descent procedure defined in~\cref{alg:latent-embedding}. In each epoch, randomly sample a subset of the training data $E\subseteq D$. We use $E$ as the data for the BLR update in~\eqref{eq:posterior}, denoting the parameters of the posterior over BLR weights as $\mathbf{m}_N^*$ and $\mathbf{S}_N^*$. Note that $\mathbf{m}_N^*$ and $\mathbf{S}_N^*$ are expressions parameterized by $\psi$. Plugging this into~\eqref{eq:predictive}, we get the moments of the predictive posterior for an arbitrary demonstration $(\mathbf{x_i},\mathbf{t_i})\in D$:
\begin{equation}
\begin{aligned}
    &\mu_N^*(\mathbf{x_i}) = {\mathbf{m}_N^*}^T\mathbf{\phi_\psi}(\mathbf{x_i}),\\
	&{\sigma^*_N}^2(\mathbf{x_i}) = \beta^{-1} + \mathbf{\phi_\psi}(\mathbf{x_i})^T \mathbf{S}_N^*\mathbf{\phi_\psi}(\mathbf{x_i}).
\end{aligned}
\label{eq:moments}
\end{equation}
The loss for our SGD is the negative log-likelihood (NLL) of $B$ for $(\mathbf{x_i}, \mathbf{t_i})$, given by:
\begin{equation}
	L(x_i, t_i) = \frac{\log(2\pi)}{2} 
	+\frac{\log({\sigma^*_N}^2(\mathbf{x_i}))}{2} + \frac{(t_i - \mu_N^*(\mathbf{x_i}))^2}{{\sigma^*_N}^2(\mathbf{x_i})}.
\label{eq:loss}
\end{equation}
In each epoch, having chosen $E$, we repeatedly sample minibatches $mb \subset D$ without replacement, and take stochastic gradient steps to minimize the NLL loss w.r.t. $\psi$.

\subsection{Reinforcement Learning with Bayesian Curiosity}
Given the latent space embedding $\phi_\psi$, we define the curiosity reward as:
\begin{equation}
    c_t =
        \log(\sigma^2(o_t)) =
        \log\left(
            \beta^{-1} + \mathbf{\phi_\psi}(o_t)^T \mathbf{S}_N\mathbf{\phi_\psi}(o_t)
        \right),
\label{eq:curiosity}
\end{equation}
where $\sigma^2$ is the variance of the predictive posterior in~\eqref{eq:predictive} and $o_t$ is the observation at time $t$. Since $\sigma^2$ is bounded from below by $\beta^{-1}$, the lower bound of $c_t$ is $- \log(\beta)$. Given extrinsic reward $e_t$, the final combined reward is:
\begin{equation}
    r_t = e_t + \eta \cdot c_t,
\label{eq:total}
\end{equation}
where $\eta$ is a hyperparameter that controls the weight of the curiosity reward.

We proceed to perform reinforcement learning with the combined reward signal $r_t$ as shown in~\cref{alg:bayesian-curiosity}. Let $A$ be some RL algorithm that can be chosen arbitrarily. Before RL begins, we reset the Bayesian linear regression $B$. In every episode, we execute rollouts of the policy $\pi_\theta$ to obtain sequences of actions, observations, and rewards. The observations $o_t$ are used to update $B$. Since $B$ is a BLR, uncertainty will, in expectation, be higher for new observations whose latent representations have low cosine similarity to previous observations. According to~\cite{ube}, states differing only in ways irrelevant to the task will be mapped to similar representations, and vice versa. Thus the curiosity reward will be higher for novel states. Note that we do not need to update $\mathbf{m}_N$, since we only use the uncertainty of $B$. This frees us from having to find target values $\mathbf{t}$, for which there is in general no intuitive candidate in the reinforcement learning setting. At the end of every episode $\pi_\theta$ is updated according to the rules of the algorithm $A$. Note that throughout the RL stage $\phi_\psi$ is not updated, as doing so would change the latent space and thus invalidate all previous updates to $B$.

\section{Experimental results}\label{results}

We proceed to test the effectiveness of our method on a series of continuous control and robotics tasks. In the following, all neural networks are implemented using Theano~\cite{theano}. Pre-training is done using ADAM~\cite{adam} with $l_2$ regularization. For the RL algorithms we use the implementations in RLLab~\cite{duan16}. Hyperparameters are $\alpha = 1e-4$ and $\beta = 1e2$ for the BLR, and the curiosity weight is $\eta = 1$. Source code is available at \url{https://gitlab.com/tomblau/Bayesian-Curiosity}.

\subsection{Classic Control}\label{toy}
\begin{table*}[tp]
\par\medskip
\renewcommand\arraystretch{1.5}
\centering
\caption{Relative improvement of using Bayesian Curiosity in classic control tasks. Showing median and interquartile range of rewards over 10 experiments, as well as relative speedup in achieving peak performance.} 
\label{table:control}
\resizebox{\linewidth}{!}{%
    \begin{tabular}{ | l | >{\centering\arraybackslash}m{1.5cm} | c | c | c | c | c | c |}
    \hline
     \multicolumn{2}{|l|}{} & \multicolumn{2}{c|}{TRPO} & \multicolumn{2}{c|}{DDPG} & \multicolumn{2}{c|}{REINFORCE}\\ \cline{3-8}
     \multicolumn{2}{|l|}{} & Reward & Speedup & Reward & Speedup & Reward & Speedup \\ \hhline{|=|=|=|=|=|=|=|=|}
     \multirow{3}{*}{Mountaincar} & Curiosity & 20.0 (16.0, 22.0) & - & -50.0 (-200.0, -7.0) & - & \textbf{30.15} (4.80, 31.70) & - \\ \cline{2-8}
     & VIME & \textbf{22.0} (17.0, 24.0) & 0.8 & - & - & 28.5 (-15.75, 30.375) & 3.36\\ \cline{2-8}
     & Vanilla & 18.0 (12.25, 22.0) & 1.43 & \textbf{11.0} (-200, 20.0) & 0.67 & 29.85 (4.75, 30.9) & 2.25\\ \hhline{|=|=|=|=|=|=|=|=|}
     \multirow{3}{*}{Swingup} & Curiosity & \textbf{-1.39} (-15.80, 12.36) & - & \textbf{33.11} (-17.78, 44.33) & - & \textbf{-46.53} (-57.61, -35.05) & - \\ \cline{2-8}
     & VIME & -46.66 (-67.93, -33.34) & 6.76 & - & - & -68.57 (-81.67, -56.04) & 5.33\\ \cline{2-8}
     & Vanilla & -23.60 (-45.99, -9.94) & 4.44 & 25.0 (-20.71, 42.22) & 1.61 & -48.09 (-57.99, -37.47) & 1.11\\ \hhline{|=|=|=|=|=|=|=|=|}
     \multirow{3}{*}{Pendulum} & Curiosity & -80.33 (-87.39, -74.47) & - & \textbf{-27.13} (-50.2, -14.43) & - & \textbf{-80.53} (-86.22, -76.41) & -\\ \cline{2-8}
     & VIME & \textbf{-79.16} (-86.03, -72.03) & 0.833 & - & - & -81.43 (-87.11, -77.19) & 2.52\\ \cline{2-8}
     & Vanilla & -79.63 (-88.25, -66.75) & 0.89 & -53.51 (-100, -19.45) & 13.33 & -83.91 (-100.0, -77.96) & 9.09\\ \hhline{|=|=|=|=|=|=|=|=|}
     \multirow{3}{*}{Acrobot} & Curiosity & \textbf{-182.7} (-241.33, -148.83) & - & \textbf{-138.05} (-203.25, -110.13) & - & \textbf{-135.5} (-190.08, -108.4) & -\\ \cline{2-8}
     & VIME & -230.1 (-319.83, -183.58) & 2.47 & - & - & -137.15 (-191.33, -110.40) & 1.14\\ \cline{2-8}
     & Vanilla & -197.8 (-286.88, 143.05) & 1.45 & -143.1 (-208.8, -111.95) & 1.52 & -145.9 (-224.13, -115.0) & 2.21\\ \hline
    \end{tabular}
}
\end{table*}

\begin{figure}[b]
\begin{center}
Mountaincar Curiosity\par\medskip
\includegraphics[width=\columnwidth]{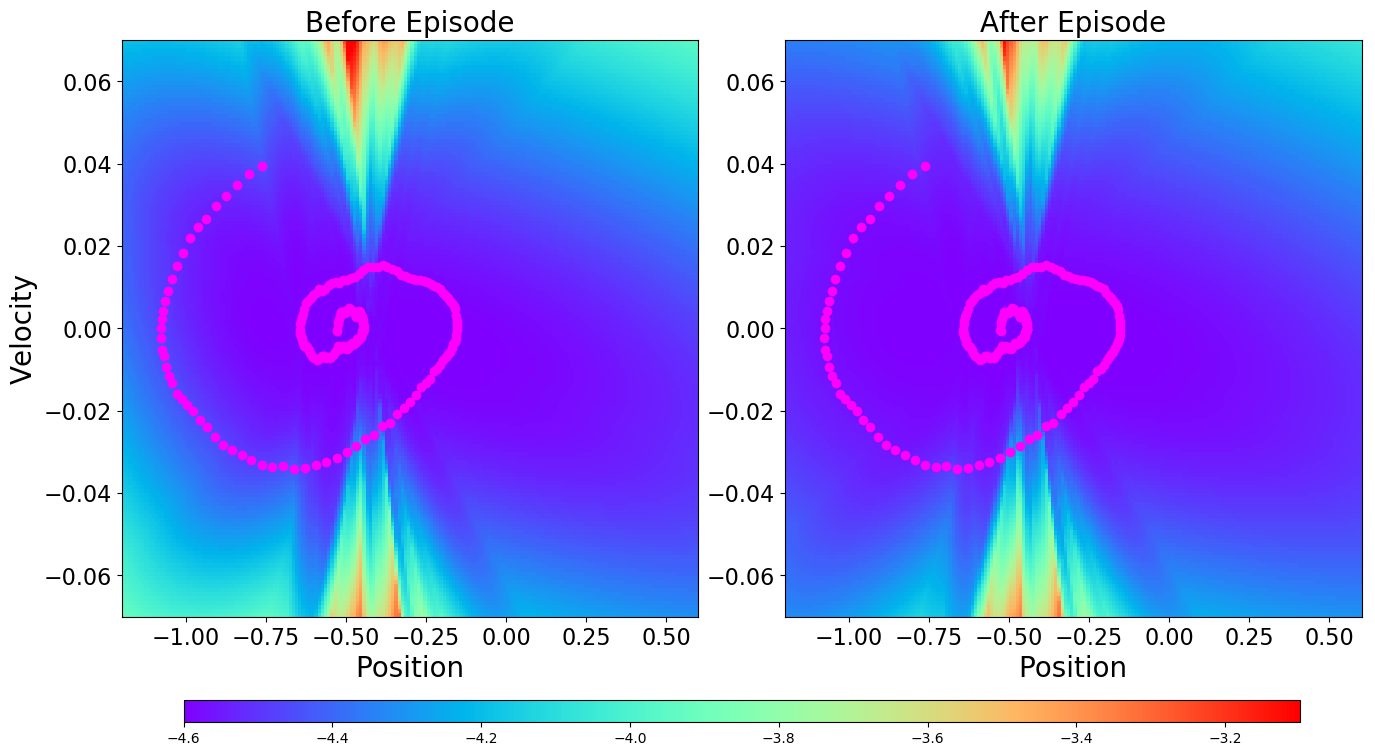}
\caption{Curiosity reward surface for mountaincar before and after an episode. Dots indicate visited states}
\label{fig:exploration}
\end{center}
\vskip -0.15in
\end{figure}

We begin by evaluating our method on classic control problems with a number of RL algorithms: REINFORCE~\cite{reinforce}, DDPG~\cite{Lillicrap15}, and TRPO~\cite{SchulmanLMJA15} in four continuous control tasks: Mountaincar, Cartpole Swingup, Pendulum, and Acrobot. For each algorithm and environment we evaluate the algorithm with and without Bayesian Curiosity. As an additional exploration baseline, we also evaluate the baseline algorithm combined with VIME~\cite{houthooft2016vime}. Extrinsic rewards for the above tasks have been sparsified to emphasize the importance of exploration.

\Cref{table:control} shows the results for this set of experiments, with each entry aggregated over 10 random seeds. The "reward" columns show the median total trajectory reward, as well as the lower and upper quartiles (the $25^{th}$ and $75^{th}$ percentiles, respectively). The "speedup" columns show how much more quickly the algorithms with Bayesian Curiosity learned compared with their respective baselines. The value is the ratio of the number of timesteps the baseline took to achieve its best result and the number of timesteps the corresponding Bayesian Curiosity algorithm took to match the baseline. Results for DDPG-VIME are missing as the authors of the original paper have not made code available for this algorithm. For most tasks and algorithms, our method is able to achieve comparable or superior performance in significantly fewer timesteps compared with both the standard RL and intrinsic motivation baselines. This is in spite of the fact that these are relatively simple problems with low-dimensional state and action spaces. Notable exceptions are DDPG on the Mountaincar task and TRPO on the Pendulum task, in which Bayesian Curiosity achieves lower performance than the baselines. In these cases, "speedup" is the ratio of the number of timesteps Bayesian Curiosity took to achieve its best result and the number of timesteps the baseline took to match it. For DDPG on Mountaincar, learning for both the baseline and Bayesian Curiosity versions is unstable, and the agents will periodically forget and re-learn how to achieve successful trajectories. Indeed, the lower quartile for both versions is $-200$ throughout the learning process, meaning that in every episode at least a quarter of agents have failed to find a successful trajectory. For TRPO on Pendulum, the baseline and Bayesian Curiosity agents achieve similar results.


Finally, we examine how the surface of curiosity rewards changes when novel states are discovered. ~\Cref{fig:exploration} shows heatmaps of curiosity rewards w.r.t. the state space before and after an episode of mountaincar. The states visited in this episode are shown as pink dots in both plots. There is a noticeable decrease in curiosity in the $\left[-1.2, -0.75\right]$ range of the x-axis, where new states were visited that have not been seen before.

\subsection{Robotic Control}
\begin{figure*}
  \par\medskip
  \centering
  \begin{subfigure}{0.33\textwidth}
    \centering
    \includegraphics[width=0.95\linewidth]{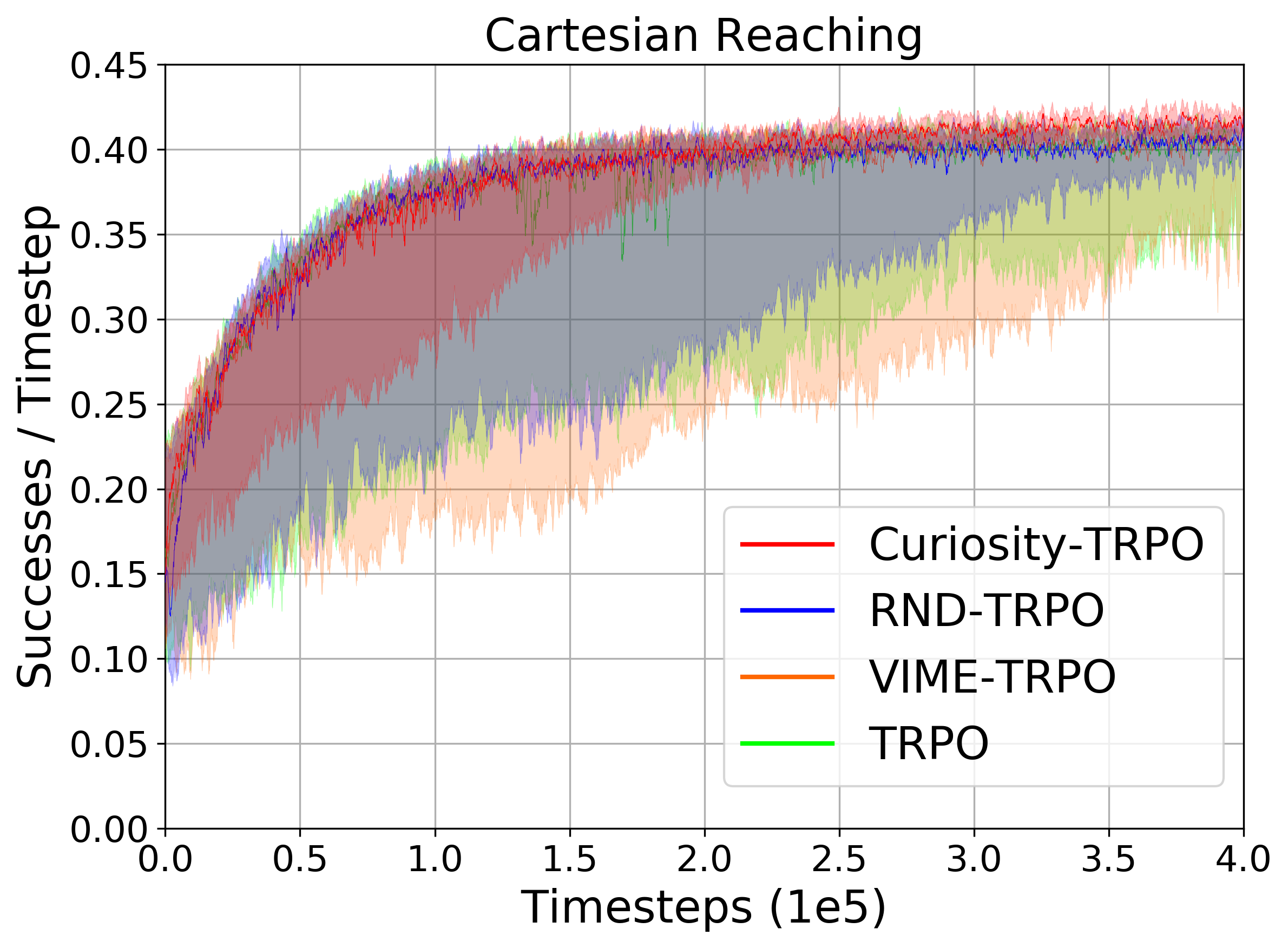}
  \end{subfigure}%
  \begin{subfigure}{0.33\textwidth}
    \centering
    \includegraphics[width=0.95\linewidth]{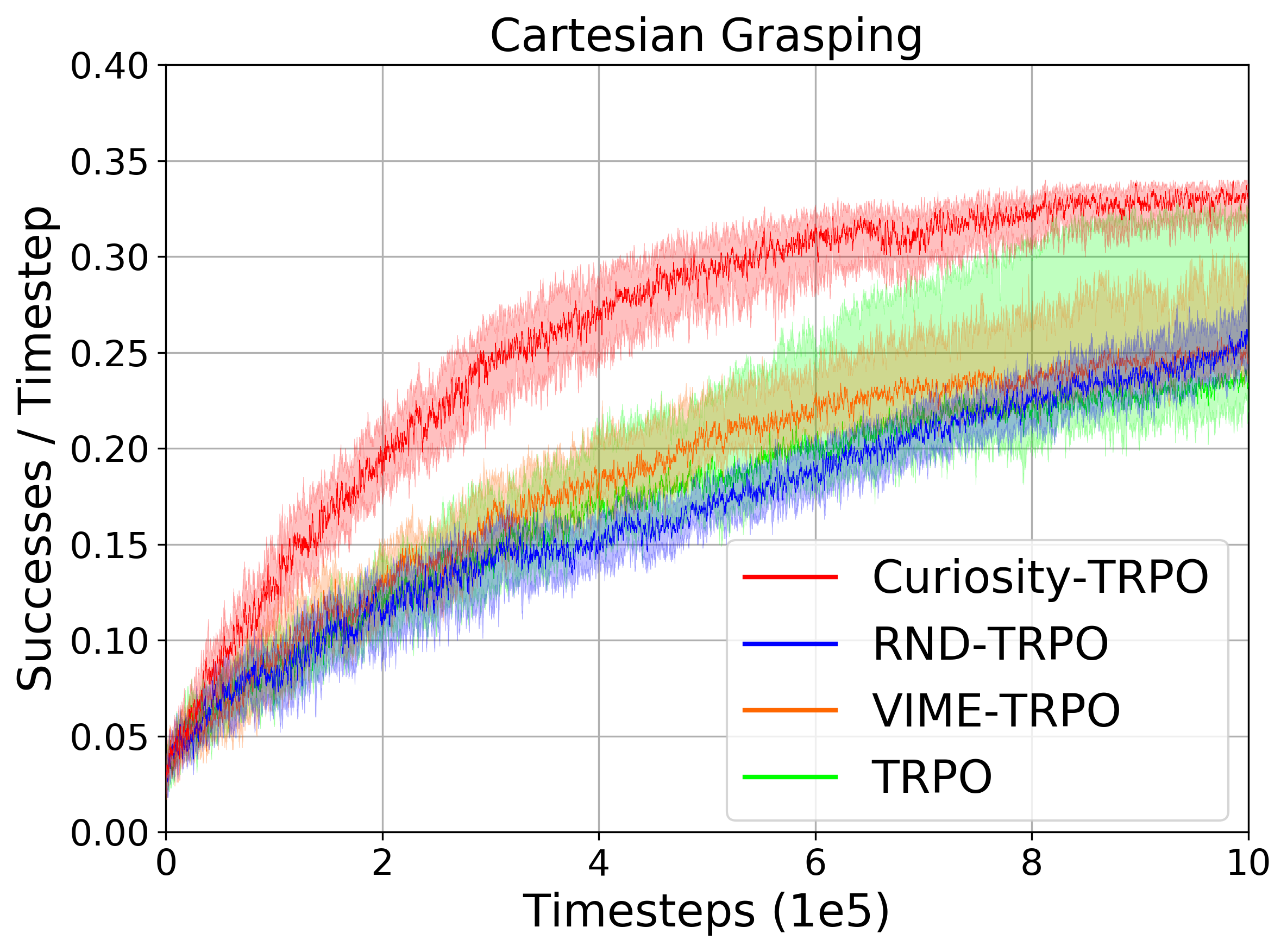}
  \end{subfigure}
  \begin{subfigure}{0.33\textwidth}
    \centering
    \includegraphics[width=0.95\linewidth]{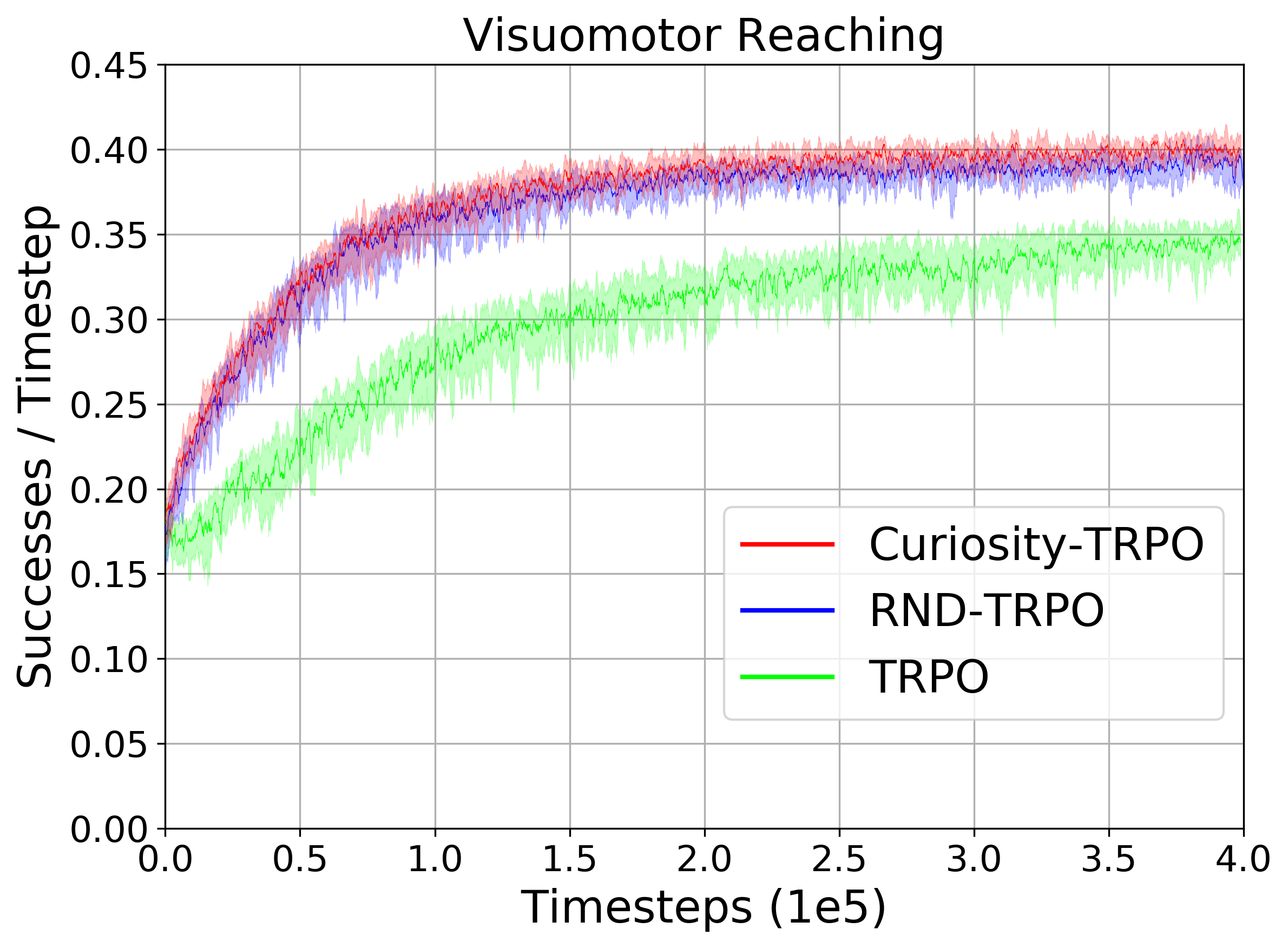}
  \end{subfigure}%
  \begin{subfigure}{0.33\textwidth}
    \centering
    \includegraphics[width=0.95\linewidth]{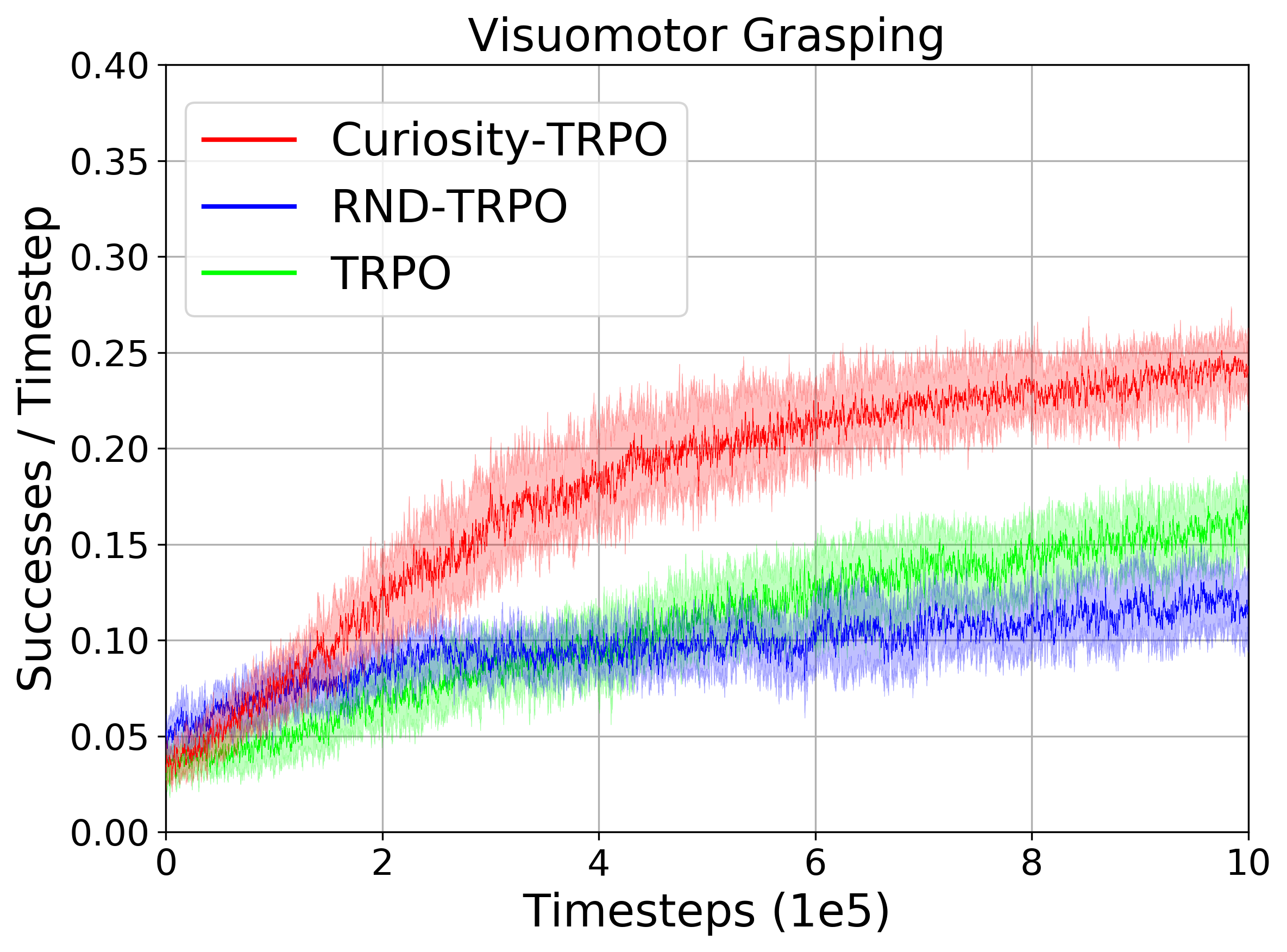}
  \end{subfigure}
  \begin{subfigure}{0.33\textwidth}
    \centering
    \includegraphics[width=0.95\linewidth]{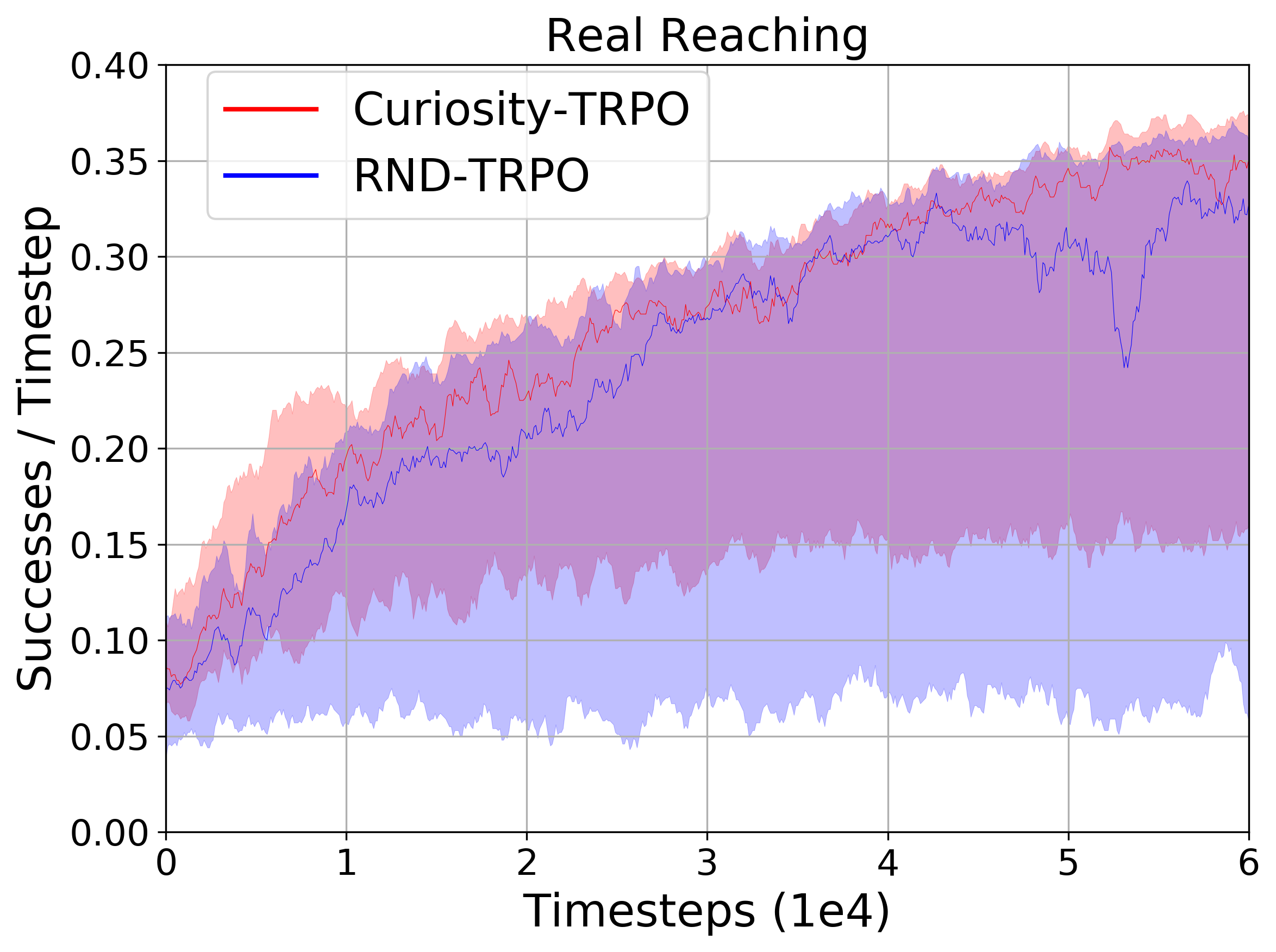}
  \end{subfigure}
  \caption{Median and interquartile range for robotic reaching and grasping, comparing TRPO with Bayesian Curiosity, with RND, with VIME, and with only extrinsic rewards. Performance is measured as the ratio of successes to timesteps. Aggregated over 5 random seeds.}
  \label{fig:robotic-control}
  \vskip -0.1in
\end{figure*}

We now examine the effect of Bayesian Curiosity rewards on learning sensorimotor control policies for a 6-DOF robotic arm. This includes \emph{Cartesian} control problems, where goal state information is represented by its $3$-d Cartesian coordinates, and \emph{visuomotor} control problems, which instead have image data. All tasks are executed in the V-REP simulation environment~\cite{vrep13} using a Kinova Jaco arm, as shown in~\cref{fig:grasp-sim}, except for the final task which is executed on a real robot. Training data for the curiosity model are generated using an inverse kinematics solver, and the set consists of $8000$ data points. To speed up the experiments and to make comparisons with the baselines more fair, all policies are pre-trained to imitate this training dataset. Experiments focus on TRPO because it was found to produce the most stable learning of all tested algorithms. All policies have Gaussian action noise determined by a neural network whose parameters are learned by the RL algorithm.

We begin with two Cartesian problems, reaching and grasping. The arm must be controlled to reach or grasp a cylinder that sits on a table. The cylinder is placed uniformly at random in a $40cm^2$ square at the beginning of an episode. In the reaching task, this cylinder is a non-interactable object that serves as an objective marker. Collision of the arm with itself or other objects results in failure. Rewards are dense for the reaching task and sparse for the grasping task. Observations consist of the angles of the $6$ controllable joints as well as the $3$-dimensional coordinates of the cylinder. Actions are a vector of angle deltas, prescribing a change in the joint angles. 

The top row of ~\cref{fig:robotic-control} shows performance on the Cartesian reaching and grasping tasks for TRPO with and without Bayesian Curiosity. We also include VIME~\cite{houthooft2016vime} and RND~\cite{burda2019exploration} as intrinsic motivation baselines. Performance is measured as $n_{succ}/n_{tsteps}$, where $n_{succ}$ is the number of successes and $n_{tsteps}$ is the number of timesteps in the 10 most recent episodes. This metric directly captures the ability of the policy to successfully complete the task, and unlike the success rate it also assigns a higher value to completing the task in fewer timesteps. In the reaching task, although median performance is similar, the lower quartile improves much more quickly in our method compared with the baselines. For the baselines, some initializations result in agents that take a very long time to converge, whereas with Bayesian Curiosity even the slowest agents converge quickly. This indicates that our method is more robust to random initialization.

For both the Cartesian and visuomotor cases, the grasping task exhibits a much starker difference between our method and the baselines than the reaching task does. The lower quartile of Bayesian Curiosity quickly rises above the \emph{upper} quartile of the baselines. Further, whereas the trendlines for the baseline agents are almost linear, the curiosity agents exhibits a more sigmoidal trendline, improving rapidly at first before slowing down. There are two factors contributing to this difference. The first is that the grasping tasks have sparse rewards while the reaching tasks have dense rewards. The second has to do with the nature of the grasping task -- any success state is very close to a failed state wherein the cylinder was knocked over rather than grasped. This means that there is a bottleneck in the state space, formed by failure states, which must be traversed to reach a success state. This kind of geometry is very difficult to explore using Gaussian action noise, and comparatively easier to explore with a curiosity reward that discourages revisiting explored states.

In the next set of experiments we investigate visuomotor versions of the reaching and grasping tasks. Observations no longer include the $3$-d coordinates of the cylinder, but instead contain RGBD images taken from a fixed camera pose.
The performance of our method is compared with the vanilla TRPO and RND baselines (VIME does not scale to high-dimensional observations) in the bottom row of~\cref{fig:robotic-control}. In the reaching task, our method greatly outperforms the naive baseline, quickly achieving performance that vanilla TRPO can't match even after $5$ times as many timesteps, and slightly improves on the RND baseline. Compared with the Cartesian case, the intrinsic motivation methods perform better, while the performance of the naive baseline degrades. This suggests that agents with intrinsic motivation can explore enough to leverage the additional information of the visual sensors, while agents that rely on Gaussian noise for exploration struggle to explore the enlarged observation space. In the grasping task, our method achieves performance that the baselines can't match even after training for $3$ times as many timesteps.

Finally, we replicate the Cartesian reaching task on a physical robotic arm. The bottom-right plot in~\cref{fig:robotic-control} shows the results of this set of experiments. Due to the time requirements of robotic experiments only RND, the strongest baseline in simulation, was included. As in the simulated reaching task, we see that median performance is comparable to the RND baseline, while the lower quartile grows more quickly, indicating higher robustness to random initialization. This robustness is particularly valuable when dealing with physical robots, as each run of the algorithm is expensive and time-consuming.

\section{Conclusions}\label{conclusions}
In this work we introduced a new method that combines BLR with a learned latent space embedding to generate a curiosity signal that directs exploration towards novel states, and demonstrated its capability to augment a variety of standard RL algorithms. Compared with both naive Gaussian noise exploration and SOTA intrinsic motivation methods, our method is able to accelerate exploration and achieve comparable or superior performance in fewer timesteps. This improvement is particularly noticeable when the state space has a geometry that makes it difficult to explore, or when the reward function is sparse, and provides little information to help direct exploration. The ability to learn policies from sparse rewards is highly desirable, as it obviates the need for carefully designing reward functions using expert knowledge.

Future work can adapt Bayesian Curiosity to continually update the latent space embedding during the RL procedure, eliminating the need for demonstrations. Another possible extension is to combine the latent space representation with Random Fourier Features, which can approximate a Gaussian Process~\cite{quia2010sparse}.

\bibliography{main}

\begin{thebibliography}{10}
\providecommand{\url}[1]{#1}
\csname url@rmstyle\endcsname
\providecommand{\newblock}{\relax}
\providecommand{\bibinfo}[2]{#2}
\providecommand\BIBentrySTDinterwordspacing{\spaceskip=0pt\relax}
\providecommand\BIBentryALTinterwordstretchfactor{4}
\providecommand\BIBentryALTinterwordspacing{\spaceskip=\fontdimen2\font plus
\BIBentryALTinterwordstretchfactor\fontdimen3\font minus
  \fontdimen4\font\relax}
\providecommand\BIBforeignlanguage[2]{{%
\expandafter\ifx\csname l@#1\endcsname\relax
\typeout{** WARNING: IEEEtran.bst: No hyphenation pattern has been}%
\typeout{** loaded for the language `#1'. Using the pattern for}%
\typeout{** the default language instead.}%
\else
\language=\csname l@#1\endcsname
\fi
#2}}

\bibitem{sutton1998reinforcement}
R.~S. Sutton and A.~G. Barto, \emph{Reinforcement Learning: An
  Introduction}.\hskip 1em plus 0.5em minus 0.4em\relax MIT press Cambridge,
  1998.

\bibitem{oudeyer2008can}
P.~Y. Oudeyer and F.~Kaplan, ``How can we define intrinsic motivation?'' in
  \emph{International Conference on Epigenetic Robotics}, 2008.

\bibitem{robbins1985some}
H.~Robbins, ``Some aspects of the sequential design of experiments,'' in
  \emph{Herbert Robbins Selected Papers}.\hskip 1em plus 0.5em minus
  0.4em\relax Springer, 1985.

\bibitem{rlsvi}
I.~Osband, B.~Van~Roy, and Z.~Wen, ``Generalization and exploration via
  randomized value functions,'' in \emph{International Conference on Machine
  Learning}, 2016.

\bibitem{2018bdqn}
K.~Azizzadenesheli, E.~Brunskill, and A.~Anandkumar, ``Efficient exploration
  through bayesian deep q-networks,'' in \emph{IEEE Information Theory and
  Applications Workshop}, 2018.

\bibitem{lipton2018bbqn}
Z.~Lipton, X.~Li, J.~Gao, L.~Li, F.~Ahmed, and L.~Deng, ``Bbq-networks:
  Efficient exploration in deep reinforcement learning for task-oriented
  dialogue systems,'' in \emph{AAAI Conference on Artificial Intelligence},
  2018.

\bibitem{ucb}
P.~Auer, ``Using confidence bounds for exploitation-exploration trade-offs,''
  \emph{Journal of Machine Learning Research}, 2002.

\bibitem{morere18a}
P.~Morere and F.~Ramos, ``Bayesian rl for goal-only rewards,'' in
  \emph{Conference on Robot Learning}, 2018.

\bibitem{ube}
B.~O'Donoghue, I.~Osband, R.~Munos, and V.~Mnih, ``The uncertainty bellman
  equation and exploration,'' in \emph{International Conference on Machine
  Learning}, 2018.

\bibitem{mnih2015human}
V.~Mnih, K.~Kavukcuoglu, D.~Silver, A.~A. Rusu, J.~Veness, B.~M. G., A.~Graves,
  M.~Riedmiller, A.~K. Fidjeland, G.~Ostrovski, \emph{et~al.}, ``Human-level
  control through deep reinforcement learning,'' \emph{Nature}, 2015.

\bibitem{thompson1933}
W.~R. Thompson, ``On the likelihood that one unknown probability exceeds
  another in view of the evidence of two samples,'' \emph{Biometrika}, 1933.

\bibitem{bellemare2016}
M.~Bellemare, S.~Srinivasan, G.~Ostrovski, T.~Schaul, D.~Saxton, and R.~Munos,
  ``Unifying count-based exploration and intrinsic motivation,'' in
  \emph{Advances in Neural Information Processing Systems}, 2016.

\bibitem{machado2018count}
M.~C. Machado, M.~G. Bellemare, and M.~Bowling, ``Count-based exploration with
  the successor representation,'' \emph{arXiv preprint}, 2018.

\bibitem{tang2017exploration}
H.~Tang, R.~Houthooft, D.~Foote, A.~Stooke, \emph{et~al.}, ``\# exploration: A
  study of count-based exploration for deep reinforcement learning,'' in
  \emph{NIPS}, 2017.

\bibitem{burda2019exploration}
Y.~Burda, H.~Edwards, A.~Storkey, and O.~Klimov, ``Exploration by random
  network distillation,'' \emph{International Conference on Learning
  Representations}, 2019.

\bibitem{pathak2017curiosity}
D.~Pathak, P.~Agrawal, A.~A. Efros, and T.~Darrell, ``Curiosity-driven
  exploration by self-supervised prediction,'' in \emph{International
  Conference on Machine Learning}, 2017.

\bibitem{achiam2017surprise}
J.~Achiam and S.~Sastry, ``Surprise-based intrinsic motivation for deep
  reinforcement learning,'' \emph{arXiv preprint}, 2017.

\bibitem{houthooft2016vime}
R.~Houthooft, X.~Chen, Y.~Duan, J.~Schulman, F.~De~Turck, and P.~Abbeel,
  ``Vime: Variational information maximizing exploration,'' in \emph{Advances
  in Neural Information Processing Systems}, 2016.

\bibitem{eysenbach2018diversit}
B.~Eysenbach, A.~Gupta, J.~Ibarz, and S.~Levine, ``Diversity is all you need:
  Learning skills without a reward function,'' in \emph{ICLR}, 2019.

\bibitem{hong2018diversity}
Z.~Hong, T.~Shann, S.~Su, Y.~Chang, \emph{et~al.}, ``Diversity-driven
  exploration strategy for deep reinforcement learning,'' in \emph{NIPS}, 2018.

\bibitem{bechtle2019curious}
S.~Bechtle, A.~Rai, Y.~Lin, L.~Righetti, and F.~Meier, ``Curious ilqr:
  Resolving uncertainty in model-based rl,'' in \emph{ICML Workshop on
  Reinforcement Learning for Real Life}, 2019.

\bibitem{yan2009linear}
X.~Yan and X.~Su, \emph{Linear Regression Analysis: Theory and
  Computing}.\hskip 1em plus 0.5em minus 0.4em\relax World Scientific, 2009.

\bibitem{Bishop2006}
C.~M. Bishop, \emph{Pattern Recognition and Machine Learning (Information
  Science and Statistics)}.\hskip 1em plus 0.5em minus 0.4em\relax
  Springer-Verlag, 2006.

\bibitem{theano}
J.~Bergstra, O.~Breuleux, F.~Bastien, P.~Lamblin, R.~Pascanu, G.~Desjardins,
  J.~Turian, D.~Warde-Farley, and Y.~Bengio, ``Theano: a {CPU} and {GPU} math
  expression compiler,'' in \emph{Python for Scientific Computing Conference
  ({SciPy})}, 2010.

\bibitem{adam}
D.~P. Kingma and J.~Ba, ``Adam: A method for stochastic optimization,''
  \emph{arXiv preprint}, 2014.

\bibitem{duan16}
Y.~Duan, X.~Chen, R.~Houthooft, J.~Schulman, and P.~Abbeel, ``Benchmarking deep
  reinforcement learning for continuous control,'' \emph{International
  Conference on Machine Learning}, 2016.

\bibitem{reinforce}
R.~S. Sutton, D.~A. McAllester, S.~P. Singh, and Y.~Mansour, ``Policy gradient
  methods for reinforcement learning with function approximation,'' in
  \emph{Advances in Neural Information Processing Systems}, 2000.

\bibitem{Lillicrap15}
T.~P. Lillicrap, J.~J. Hunt, A.~Pritzel, N.~Heess, T.~Erez, Y.~Tassa,
  D.~Silver, and D.~Wierstra, ``Continuous control with deep reinforcement
  learning,'' \emph{arXiv preprint}, 2015.

\bibitem{SchulmanLMJA15}
J.~Schulman, S.~Levine, P.~Moritz, M.~I. Jordan, and P.~Abbeel, ``Trust region
  policy optimization,'' \emph{International Conference on Machine Learning},
  2015.

\bibitem{vrep13}
E.~Rohmer, S.~P.~N. Singh, and M.~Freese, ``V-rep: a versatile and scalable
  robot simulation framework,'' in \emph{International Conference on
  Intelligent Robots and Systems}, 2013.

\bibitem{quia2010sparse}
J.~Quinonero-Candela, C.~E. Rasmussen, A.~R. Figueiras-Vidal, \emph{et~al.},
  ``Sparse spectrum gaussian process regression,'' \emph{Journal of Machine
  Learning Research}, vol.~11, 2010.

\end{thebibliography}
\bibliographystyle{IEEEtran}

\end{document}